# Image Segmentation Based on Histogram of Depth and an Application in Driver Distraction Detection

Tran Hiep Dinh, Minh Trien Pham, Manh Duong Phung, Duc Manh Nguyen, Van Manh Hoang, Quang Vinh Tran
University of Engineering and Technology (UET)
Vietnam National University, Hanoi (VNU)
Hanoi, Vietnam
tranhiep.dinh@vnu.edu.vn

***Abstract*—This study proposes an approach to segment human object from a depth image based on histogram of depth values. The region of interest is first extracted based on a predefined threshold for histogram regions. A region growing process is then employed to separate multiple human bodies with the same depth interval. Our contribution is the identification of an adaptive growth threshold based on the detected histogram region. To demonstrate the effectiveness of the proposed method, an application in driver distraction detection was introduced. After successfully extracting the driver's position inside the car, we came up with a simple solution to track the driver motion. With the analysis of the difference between initial and current frame, a change of cluster position or depth value in the interested region, which cross the preset threshold, is considered as a distracted activity. The experiment results demonstrated the success of the algorithm in detecting typical distracted driving activities such as using phone for calling or texting, adjusting internal devices and drinking in real time.**

## I. INTRODUCTION

Human segmentation and recognition from images is today's key challenge in computer vision. With the release of Kinect camera [1], researchers could be able to develop new solutions based on depth information. Several approaches were comprehensively reviewed in [2]. Spinello in [3] introduced a combo detector based on histogram of oriented depths and histogram of oriented gradients in RGB data. The approach was demonstrated to detect several people at different ranges in different visual clutter conditions. A similar idea was also presented by Choi by fusing together multiple image-based and depth-based methods such as pedestrian and upper body, face, skin, shape and motion detectors [4].

Human body could also be segmented from the depth images using background subtraction algorithm [5][6]. A depth background image was obtained from a training set. A pixel of current frame is considered as foreground if its difference between corresponding background pixel cross the predefined threshold, which was calculated based on the mean value and standard deviation for each pixel.

Lu Xia proposed a model based approach to segment the human from the surroundings by using 2D chamfer distance matching and 3D model fitting [7]. The contour of segmented body was then extracted accurately by the region growing algorithm, which was again applied in our study with modifications to obtain more robust results.

Exploiting the fact that the depth values representing the interested object should vary smoothly in a specific range, we employ in this paper a histogram of depth to extract and separate objects from its surroundings. Since the histogram describes the repeated times, not the space distribution of depth values, it could only segment a group of objects that have the same range of depth. In order to overcome this, a region growing algorithm is applied to determine the bounding boxes of potential objects. The scale of the bounding boxes is then used to identify the human body. The proposed method was applied to a distracted driving detection application in which the driver position is extracted in the first frame and used as reference. A comparison between the present and reference frames provides information to analyze the change in motion and depth values.

The paper is structured as follows. The histogram-based segmentation is introduced in section 2. Section 3 describes the application of distracted driving detection. Experiments and conclusions are presented in the last two sections.

## II. HISTOGRAM BASED SEGMENTATION

### A. Regions of Interest (ROI) Detection

The depth image is analyzed to determine the appeared depth values and its corresponding number of repetition. The data noise, which is represented as zero value, will be ignored. Let $X$ be a vector of ascending sorted depth values $X = [x_1, x_2, ..., x_n]$ $(x_1 < x_2 < ... < x_n)$ and $Y$ be a vector of corresponding repetition of each measurement $Y = [y_1, y_2, ..., y_n]$. An element $Y_i$ is considered as peak if its value is greater than its two neighbors:

$$\begin{cases} Y_i > Y_{i-1} \\ Y_i > Y_{i+1} \end{cases} \quad (1)$$

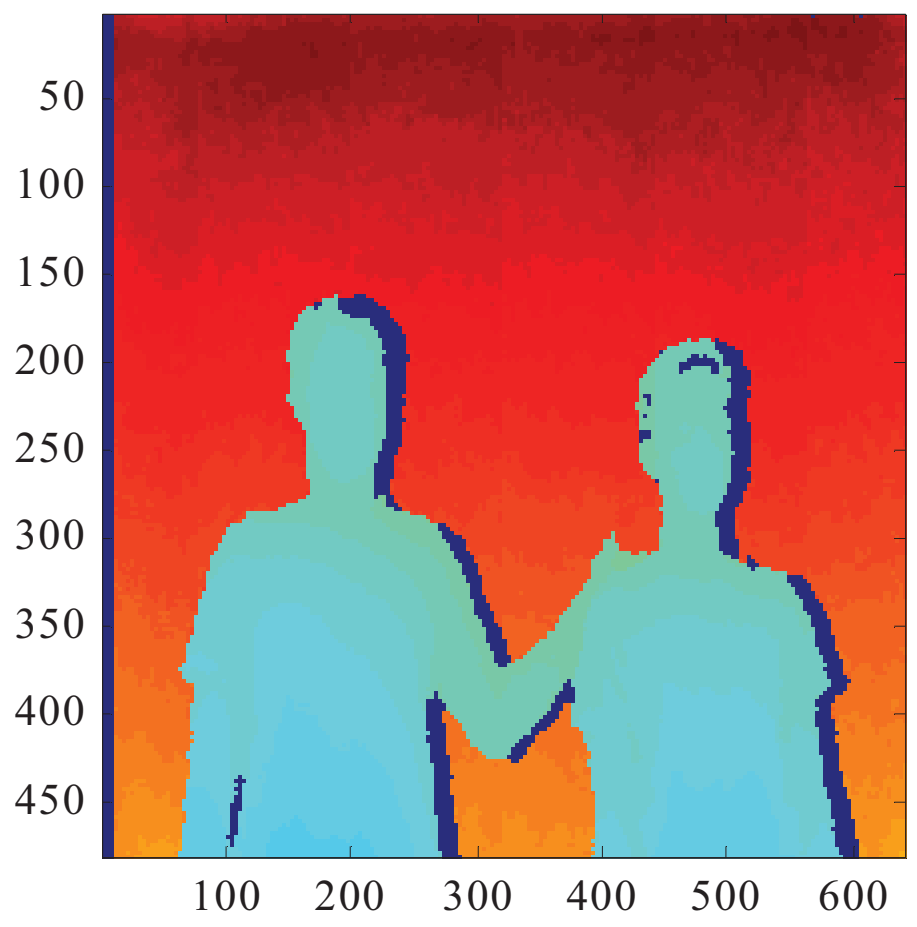

(a)

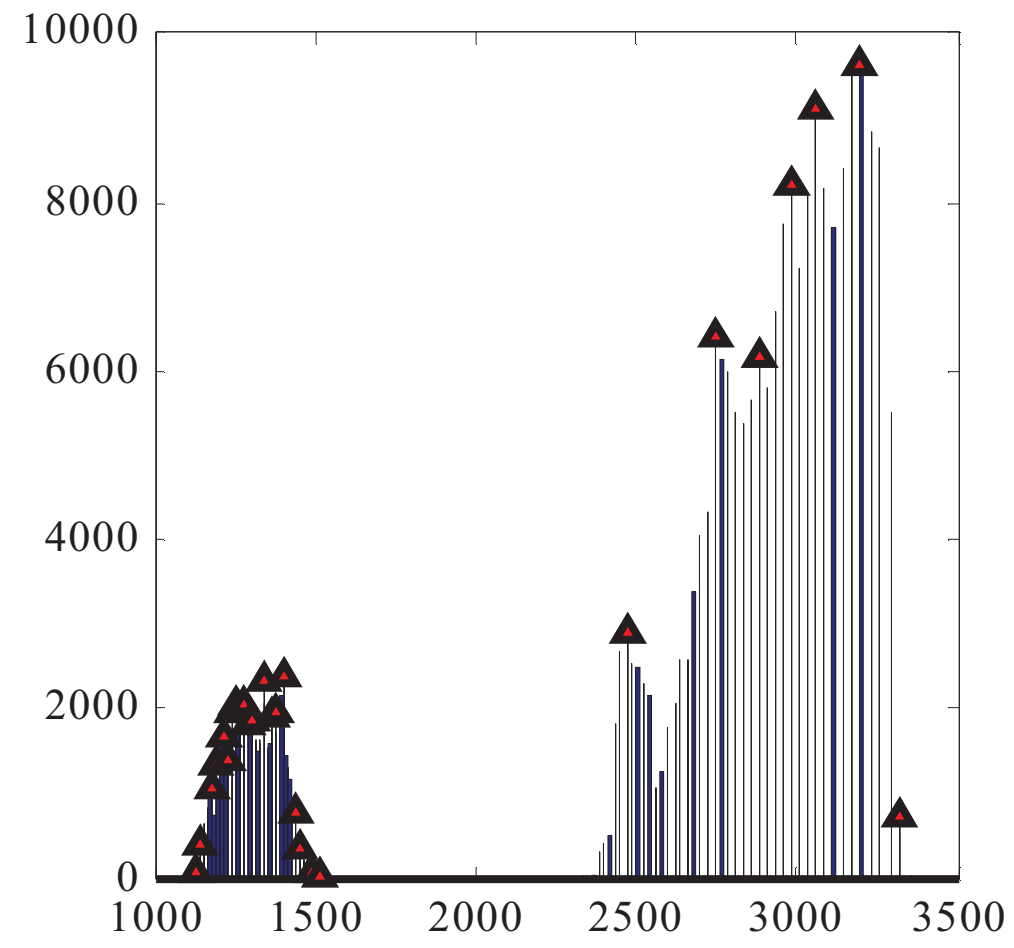

(b)

Figure 1. Histogram of depth and peaks detection

(a) Depth image of sample scene

(b) its corresponding histogram with peaks

Fig. 1a illustrates a scene with two men standing before a flat background. Fig. 1b presents the histogram of depth and detected peaks. The peaks number and distribution of depth values at this stage provide noisy information to identify the regions of interest. The histogram shape of a human body also varies based on his poses. Therefore, using local peaks to segment image could be ineffective with depth data.

Let $T$ be the difference between the nearest and farthest distance of human body (commonly $T = 40cm$ ) and $P = [p_1, p_2, ..., p_k]$ ($k < n$) be the vector containing detected peaks. The algorithm to segment the obtained depth map is as follows:

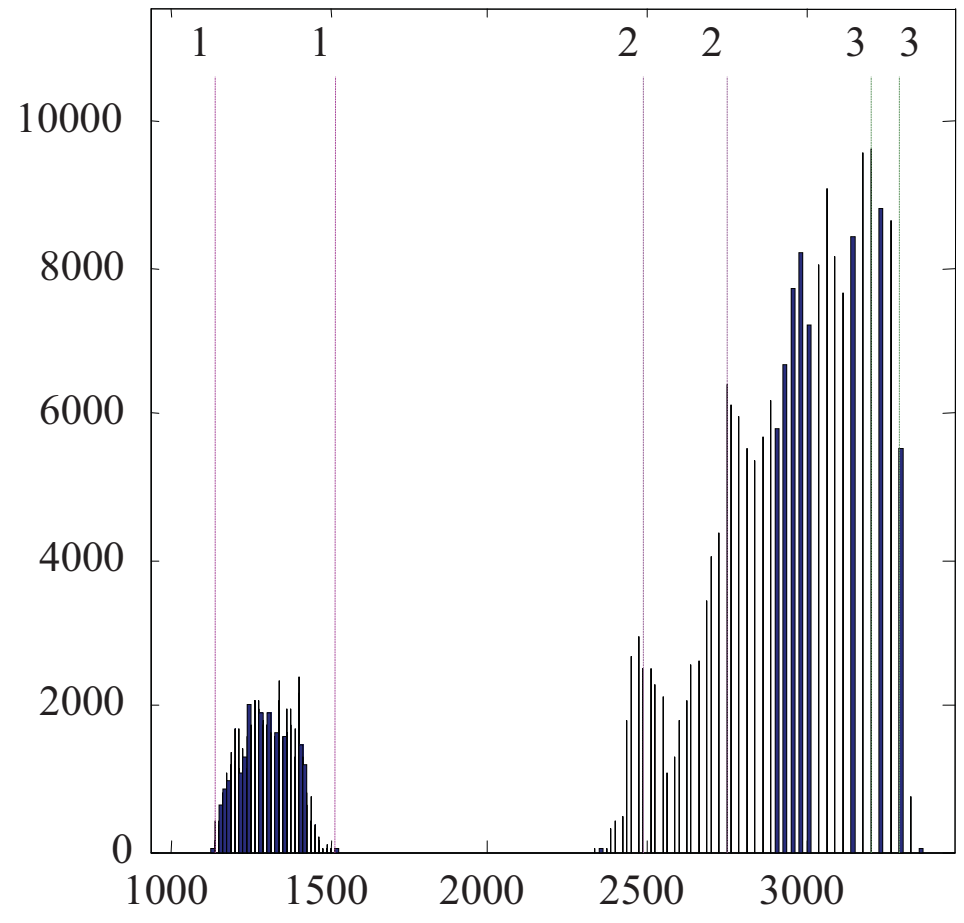

(a)

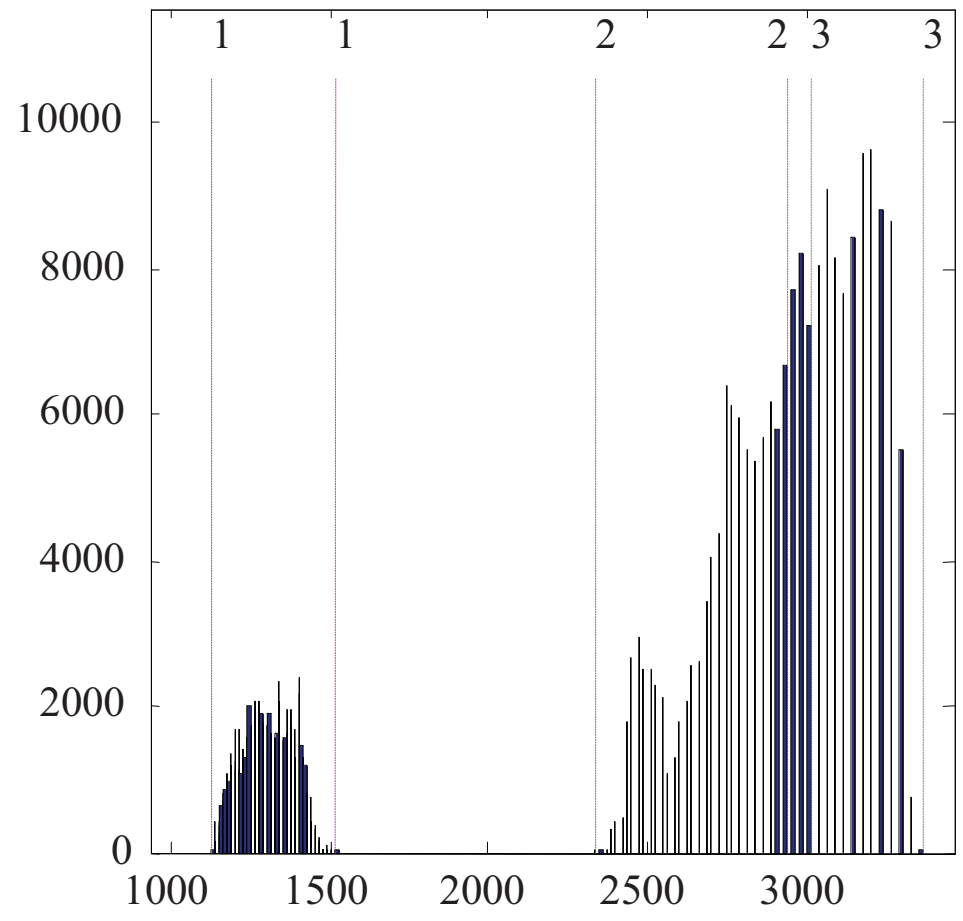

(b)

Figure 2. Segmentation process based on HOD

(a) A Histogram with three ROI

(b) The ROI are made wider

For i=1:k

if

$$\left| X(p_{i+1}) - X(p_i) \right| > T \tag{2}$$

(i) split the peaks array in to two parts $P_1 = [p_1, p_2, ... p_i]$ and $P_2 = [p_{i+1}, p_{i+2}, ... p_k]$

(ii) for each new part, compute the new peaks as in (1) to reduce the unexpected peaks

The split and reduce process is repeated on partial array until only one peak left or the width of the remained region smaller than the defined threshold $T$:

$$\begin{cases} length(P) < 2 \\ \left| P(first) - P(last) \right| < T \end{cases} \tag{3}$$

Fig. 2a shows the segmentation process in which the histogram is divided into three regions. To avoid the loss of information, the detected region is made wider from both sides for $T/2$ before being narrowed down to its nearest valley as shown in fig. 2b.

### B. *Adaptive Region Growing Algorithm*

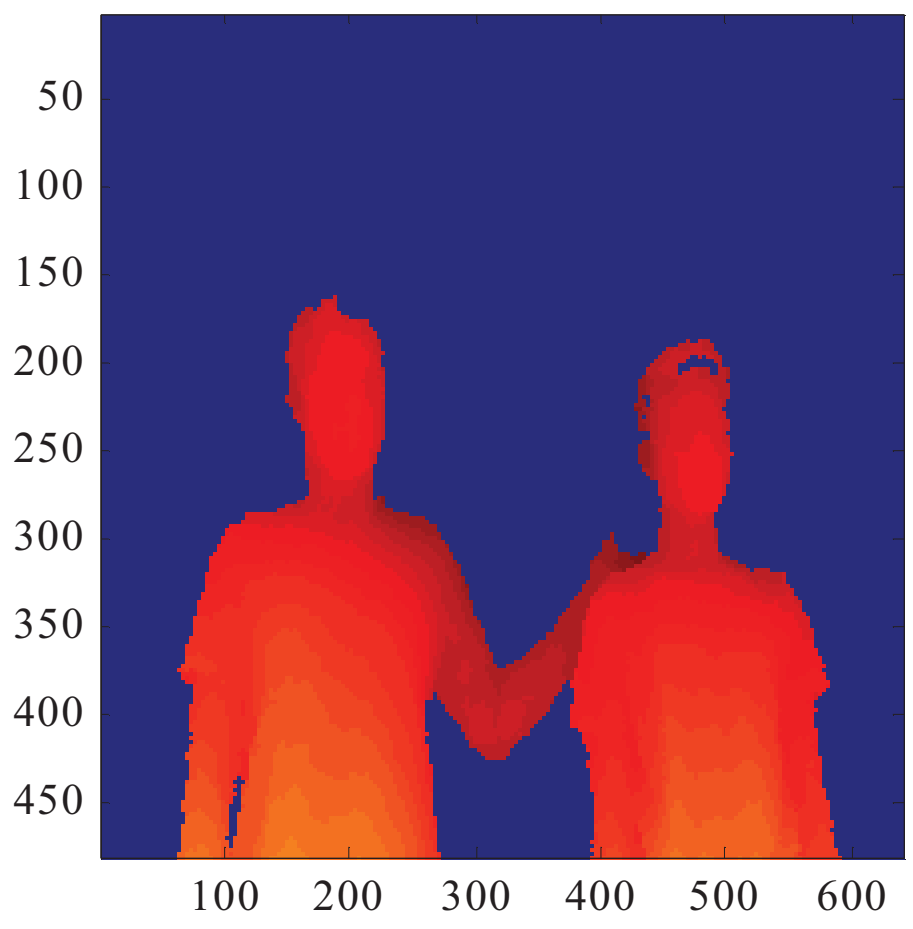

(a)

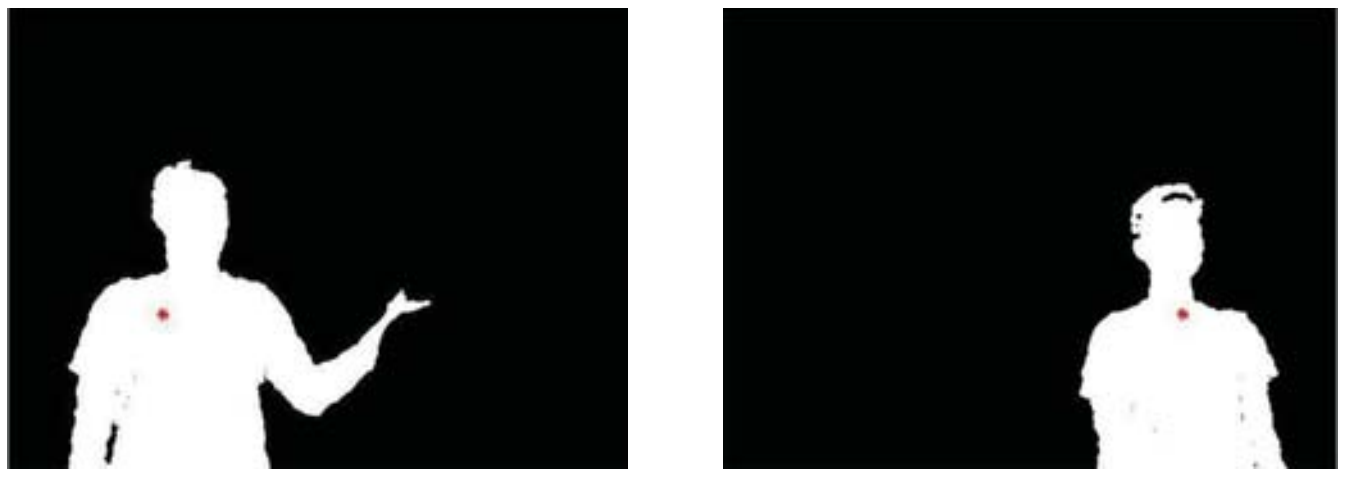

(b)

Figure 2. Performance of region growing algorithm on two objects with a same interval of depth value

(a) ROI extracted by peaks detection

(b) The final extraction result using seed points and adaptive threshold

On each detected region, a region growing process is applied to separate objects that have the same interval of depth. Similarly to [7], main factors of the growing process algorithm are the seed location and the growth threshold. Each detected region is scanned vertically and horizontally to locate the position that have the highest density of depth value. This position has the high potential to be an interested object and thus chosen as seed points. The similarity $S(x, y)$ between two pixels $x$ and $y$ are defined alike in [7]:

$$S(x, y) = \left| depth(x) - depth(y) \right| \quad (4)$$

The threshold is determined based on the histogram of depth (HoD) of detected regions. Unlike [7], a fixed growth threshold is not set in this study due to the fact that the density of depth points on each object could be different and consequently could lead to misidentify if the interested object changes its position. Instead, for each found seed point $SP$, the location of corresponding bar $X_{SP}$ in the HoD is determined. The threshold for adding neighbor pixels of each seed $S$ is then defined by comparing the depth of $SP$ to its neighbors:

$$T_H = \max(X_{SP+1} - X_{SP}, X_{SP} - X_{SP-1}) \quad (5)$$

As the depth value of each seed varies with its position, the threshold $T_H$ is not fixed but adapts accordantly. As the result, this region growing process ensures that all added neighbor pixels belong to two consecutive existing depth layers of the detected region. The efficiency of the process is enhanced. We call it *adaptive region growing algorithm*.

Fig. 3a shows a region of interest extracted by the segmentation process II.A in which two people are not separated. Fig. 3b shows the result of seed point detection and advanced region growing algorithm. Thanks to the adaptive growth threshold, each person was correctly extracted.

## III. An Application in Driver Distraction Detection

Distracted driving could increase crash risks. Using in-vehicle technology to detect distraction is a pressing concern in recent years. Sean proposed an approach using the skeletal tracking system of Kinect to estimate distance between specific joints [8]. The study is, however, limited by misidentified joints when the drivers move quickly. In this work, we address typical distracted activities during driving such as drinking, using a handheld cell phone and manipulating vehicle instruments [9]. The approach consists of two steps. First, the driver's position is located by the algorithm presented in Section II. The distraction is then monitored by a motion tracking process. As this novel approach uses only the depth images and compact computation, it archives real-time performance while maintains the accuracy rate of 100% among all our experiments.

### A. *Driver Location Identification*

The segmentation algorithm in section II is employed to detect the driver position and motion. First, the depth map of in-car environment is segmented. The shape of the extracted region is then evaluated using the bounding box. Its size and ratio between the length and width are factors to identify the driver location. As the driver will stay on his driver's seat, only the change of depth inside the corresponding bounding box is analyzed to detect the driver's motion. Limiting the process inside the bounding box not only eliminates noises but also reduces the computing time without affecting the results.

### B. *Motion Tracking*

The motion information is extracted by comparing the variation of depth value between the initial and present frames. A similar idea was mentioned in [10] but the difference between each pair of consecutive depth frames was concerned. In our work, we assume that the driver's pose in the initial frame is the safe position for driving. Thus, when tracking motion by subtracting the initial from the current depth frame, the depth change provides more information for evaluation. More specifically, we analyze the size and depth value of the adjustment sectors, while the driver is posing some typical distracted activities, with hand off the wheel.

Let $A_R$ be the total number of pixels belonging to the region of driver in the reference frame. Connected component algorithm is applied to the subtracting result between the current and initial frame to identify areas of which the pixels values are greater than null. The relative size $A_{changed}$ of these areas is calculated by dividing its number of pixels $A_C$ by $A_R$ as in (6). The mean depth $D_{changed}$ of each area is also calculated as in (7):

$$A_{changed} = \frac{A_C}{A_R} 100\% \tag{6}$$

$$D_{changed} = \frac{\sum_{i=1}^{A_C} depth(x_i, y_i)}{A_C} \tag{7}$$

For each investigated frame, the total change in size and the mean of change in depth are recorded to issue warning messages for the driver's distracted activities.

## IV. EXPERIMENT RESULTS

We used a Kinect camera to record video sequences for test scenarios. For in-vehicle experiments only, the camera was set to near mode in order to avoid losing information when the driver's hand moving too close to the Kinect. Other studies were taken with Kinect's default range. The algorithm was implemented in MATLAB.

### A. Object Segmentation

In the first experiment, 4 persons appeared with difference distances to the camera (fig. 3a). The regions segmented by using HoD are shown in fig. 3b. The nearest person was detected in region 1 while others appeared in region 2. Fig. 3c shows the result of the separating process of foreground objects. All parts belonging to participated members were successfully extracted. The segmentation results were still correct when they changed their pose or position.

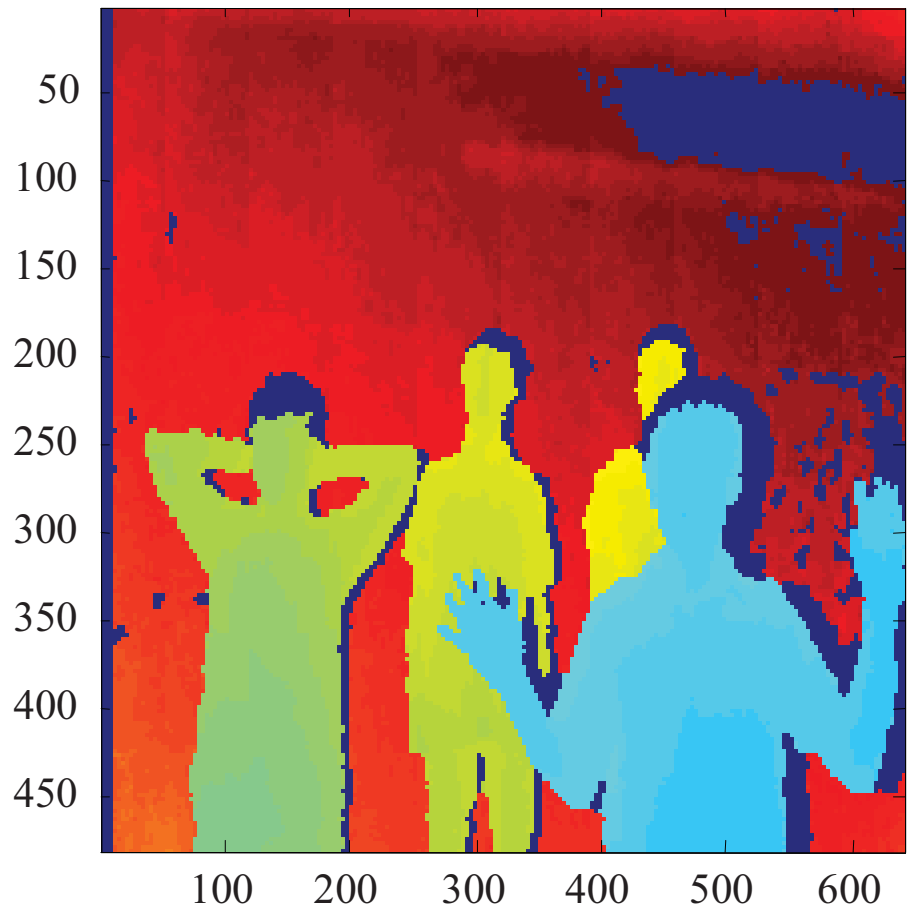

(a)

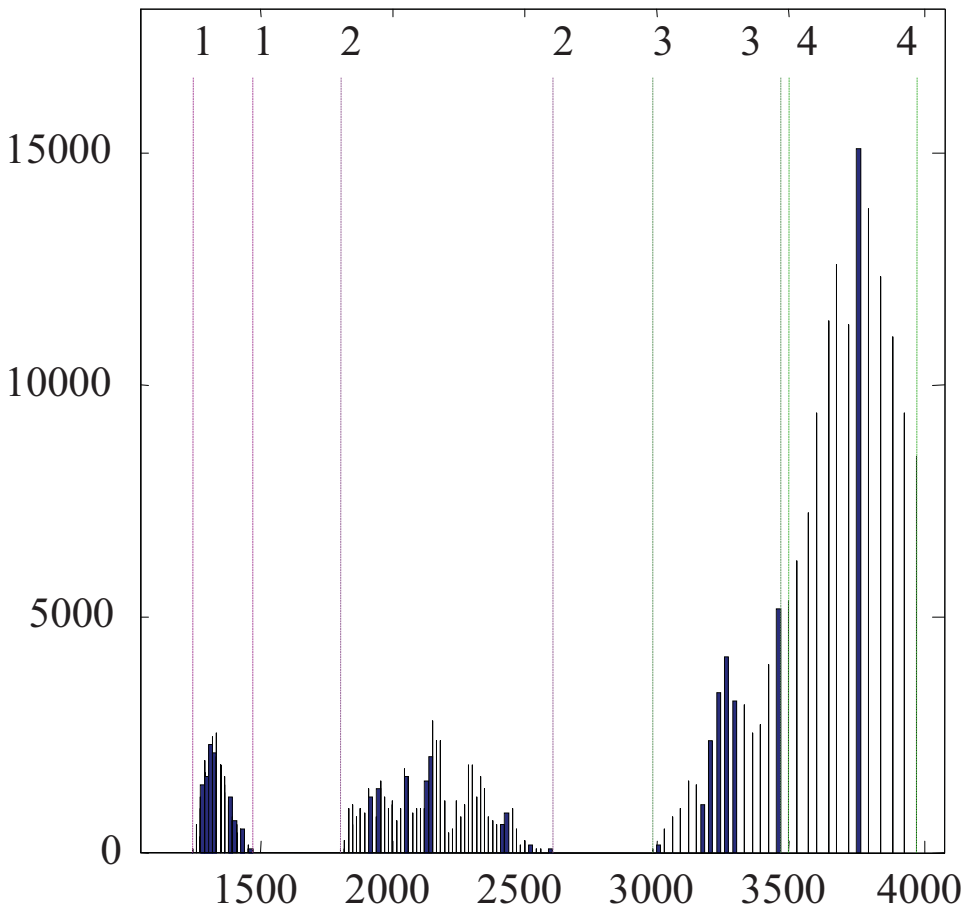

(b)

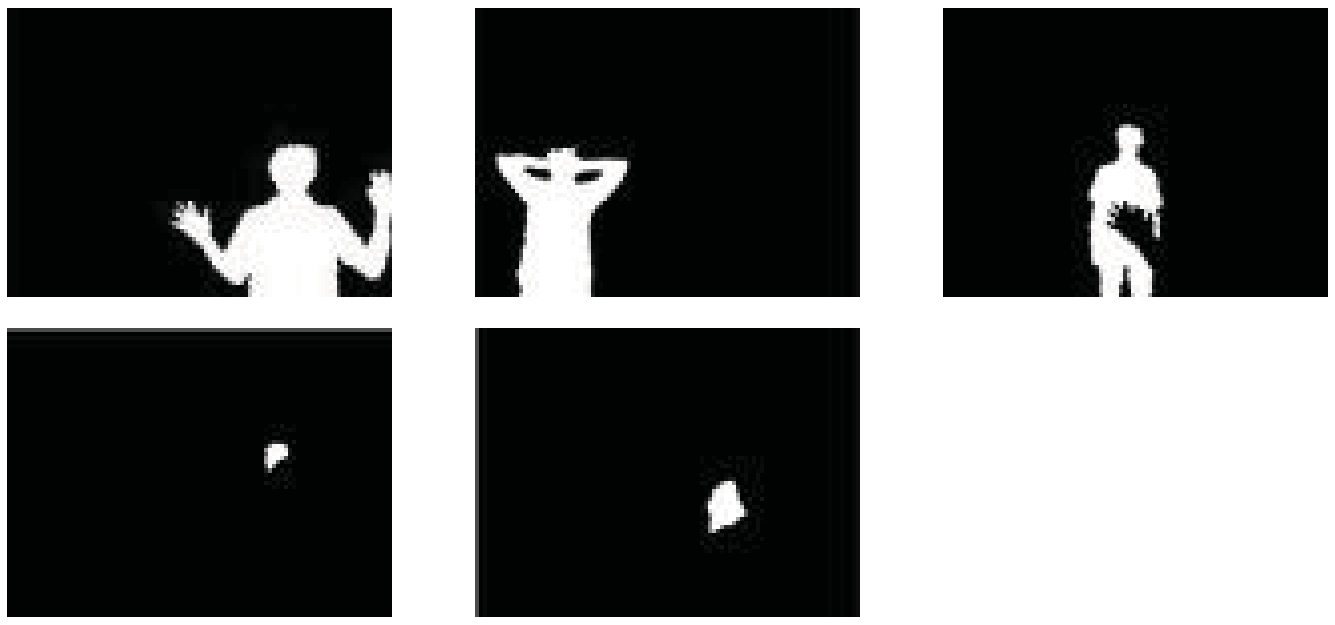

(c)

Figure 3. Test scene with 4 men

In the second experiment, the Kinect camera was mounted inside a car and directed to the driver. Two different cars and drivers were involved. Fig. 4a shows the original depth image and the segmented driver pose for the 4-seat car. The result for the 7-seat car is shown in fig. 4b.

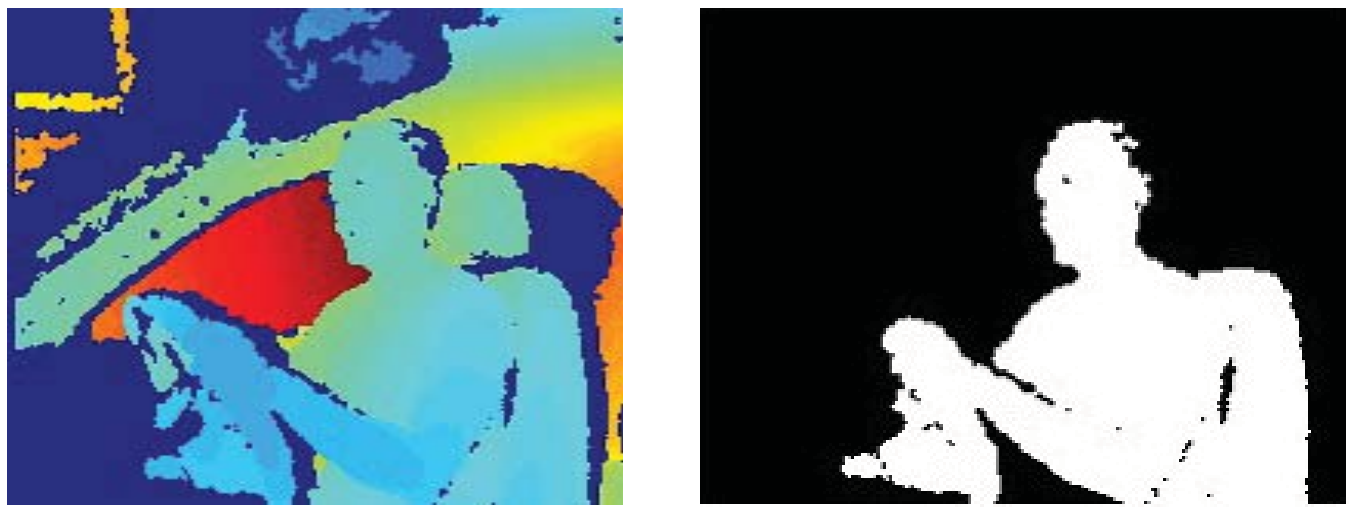

(a)

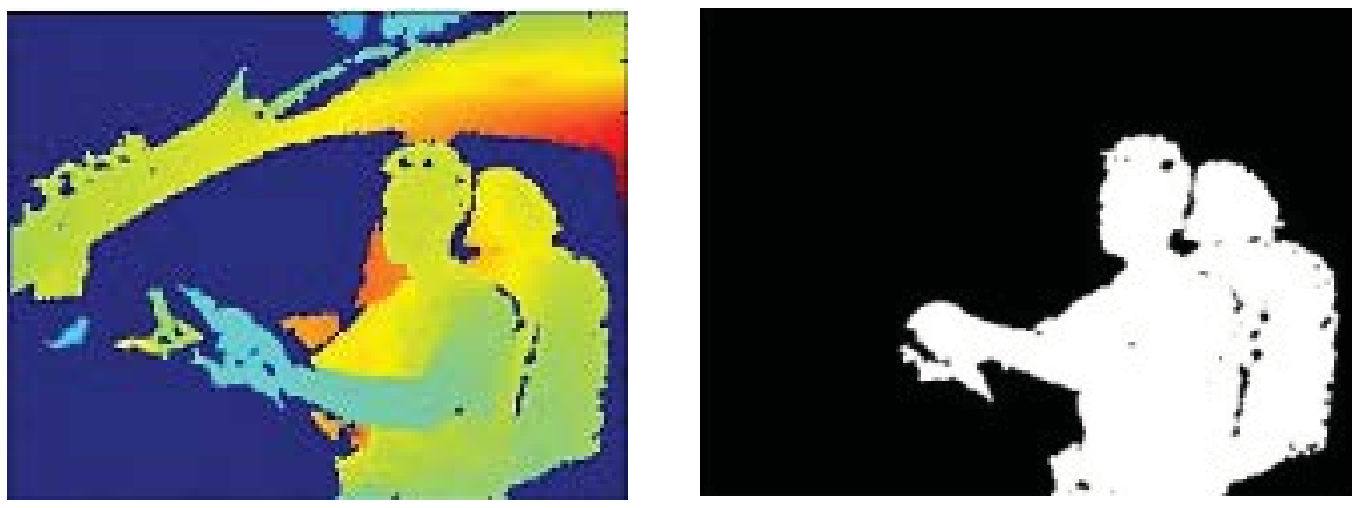

(b)

Figure 4. Driver position identification

(a) 4 seats car

(b) 7 seats car

## B. *Depth Change Analysis*

### 1) *Image Subtraction*

In this experiment, the drivers were required to perform actions related to distracted driving activities such as texting, calling, drinking and manipulating vehicle instruments. The variation of depth between the present and reference frames was analyzed to determine the path of motion. The values of depth change were also taken into account as shown in fig. 5a in which the color from dark to light corresponding to the change from major to minor.

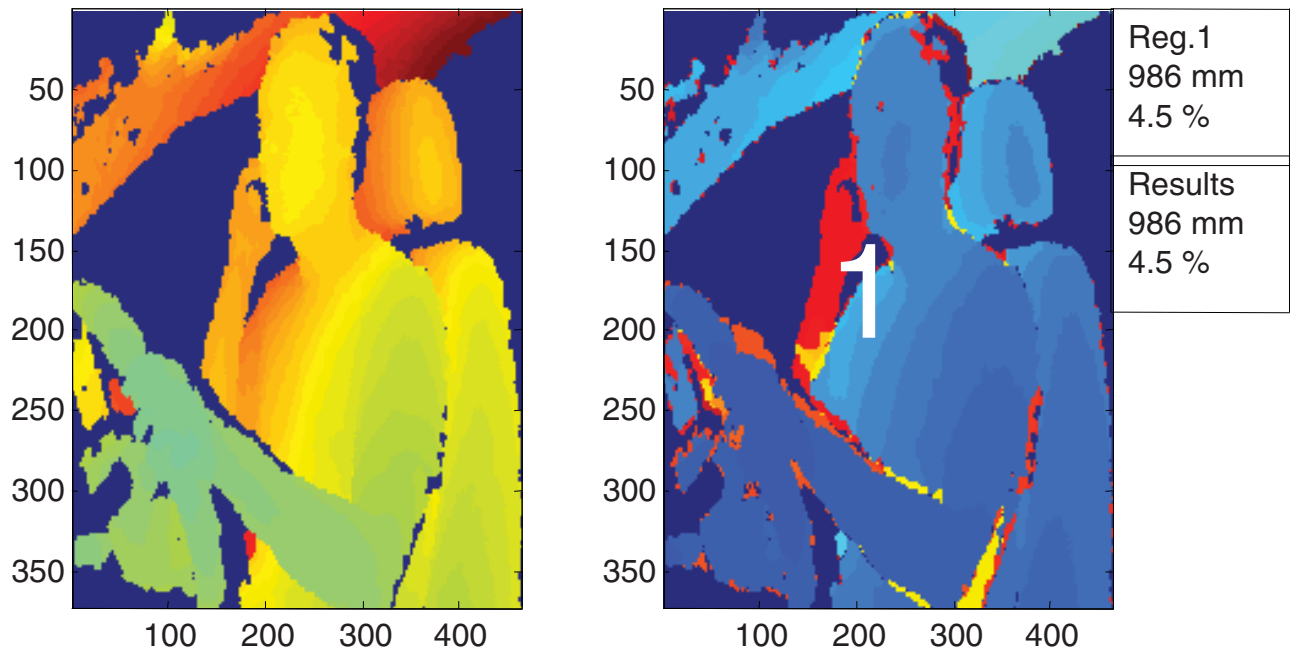

(a) (b)

Figure 5. An example of image subtraction

(Calling while driving)

(a) Extracted depth image

(b) Difference depth image

The size of the depth image in this case is 370x460 instead of 480x640 as in the original image as we only track the change inside the bounding box of the extracted driver object. For better explanation, we recalibrated the difference image into a gray scale image in which the unchanged pixels took value from 0 to 100 and changed pixels was in the range from 150 to 255 as shown in fig. 5b.

### 2) *Noise Removing*

As can be seen in fig. 5b, some red points appeared at locations such as the driver seat and car ceiling where changes were not expected to occur. These noises could affect the evaluation result. In this paper, this problem is partially solved by representing the data in a lower resolution grid map where each cell receives the highest value among the depth pixels inside. The result of the motion analyzing in lower resolution was illustrated in fig. 6.

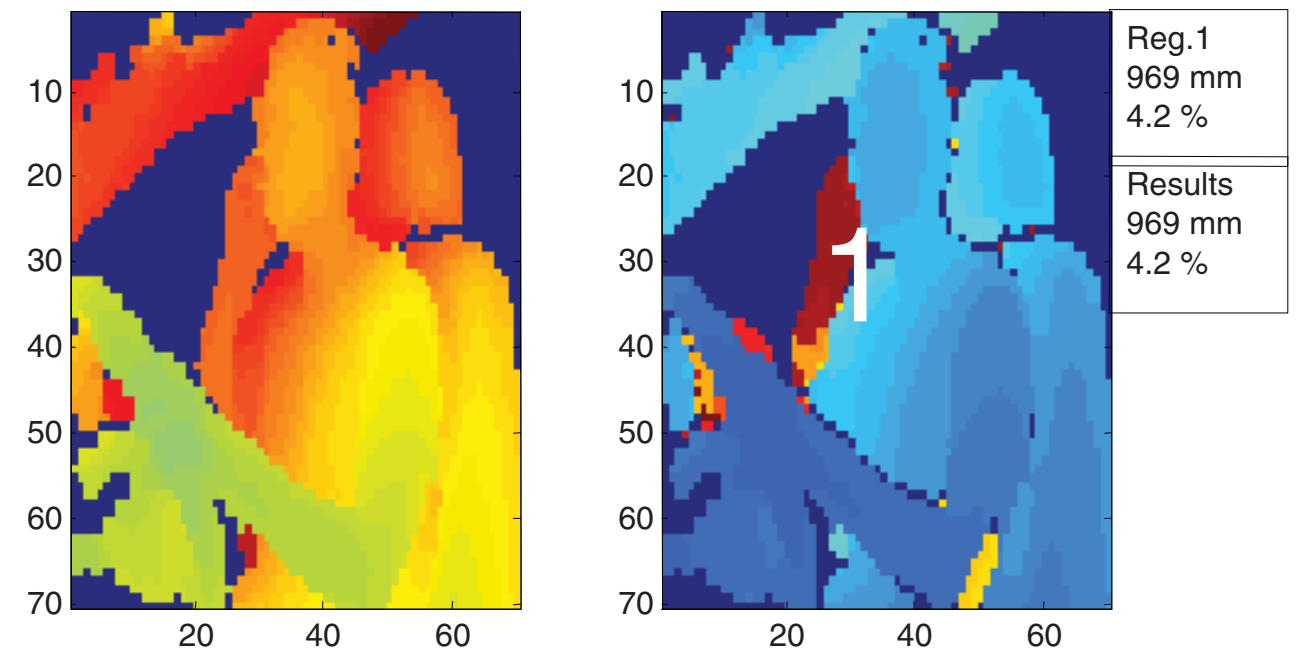

(a) (b)

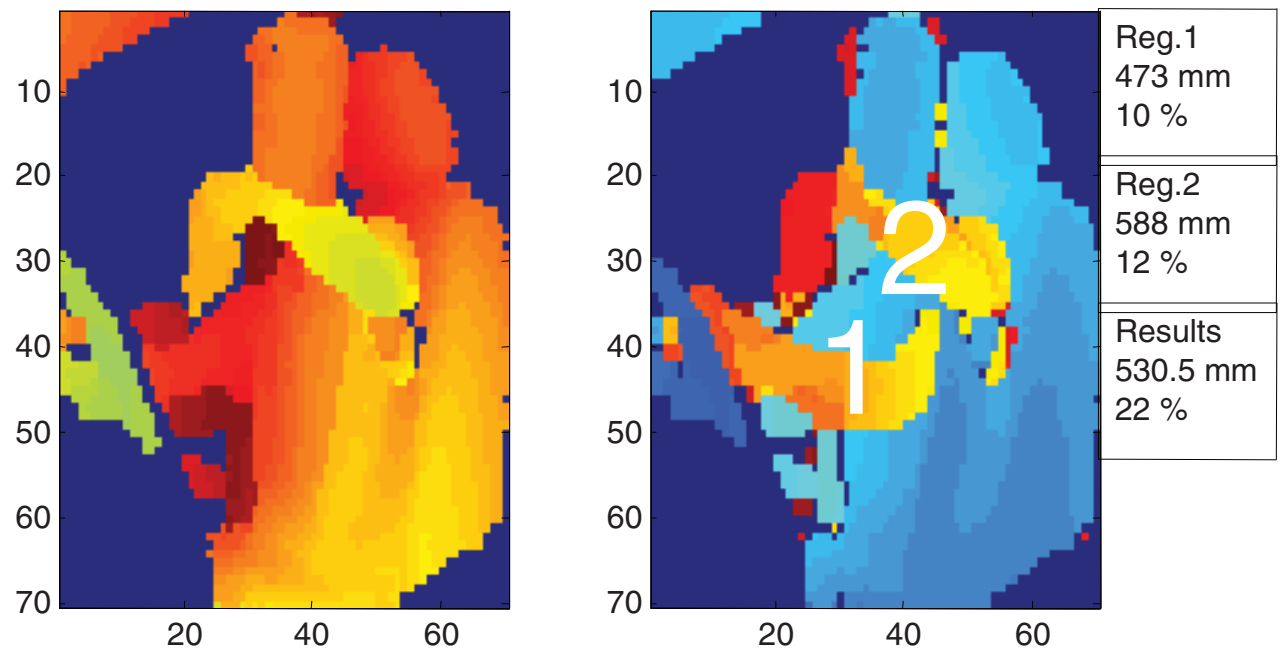

(c) (d)

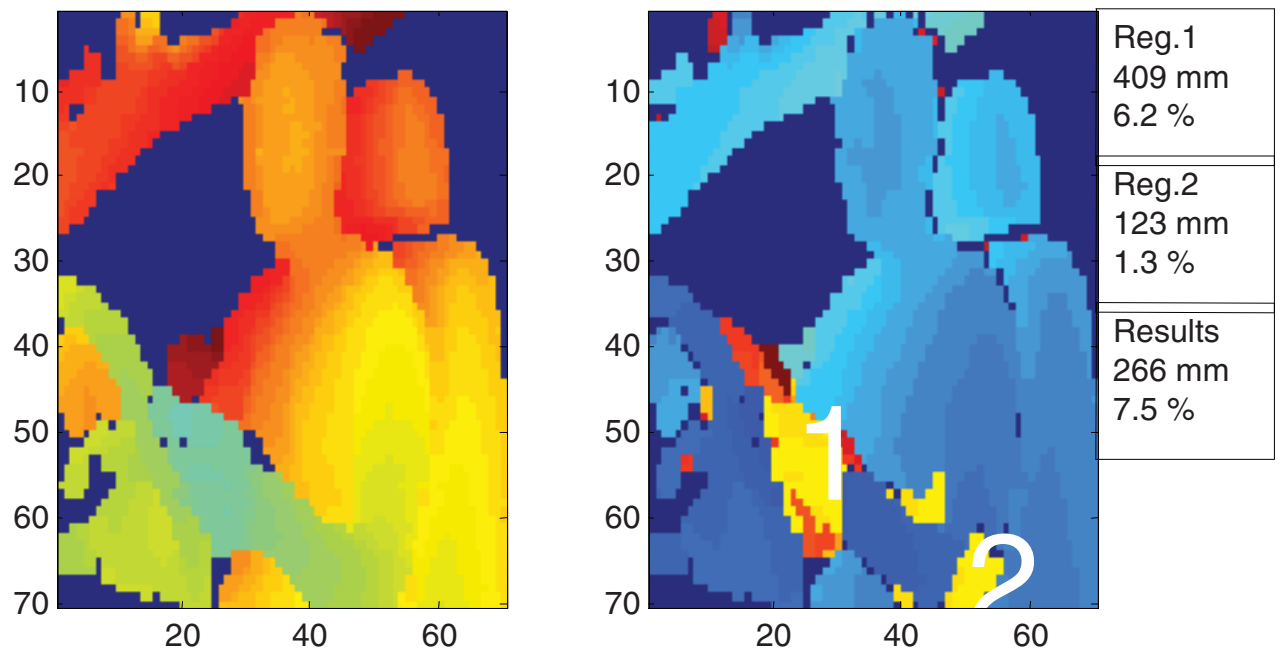

(e) (f)

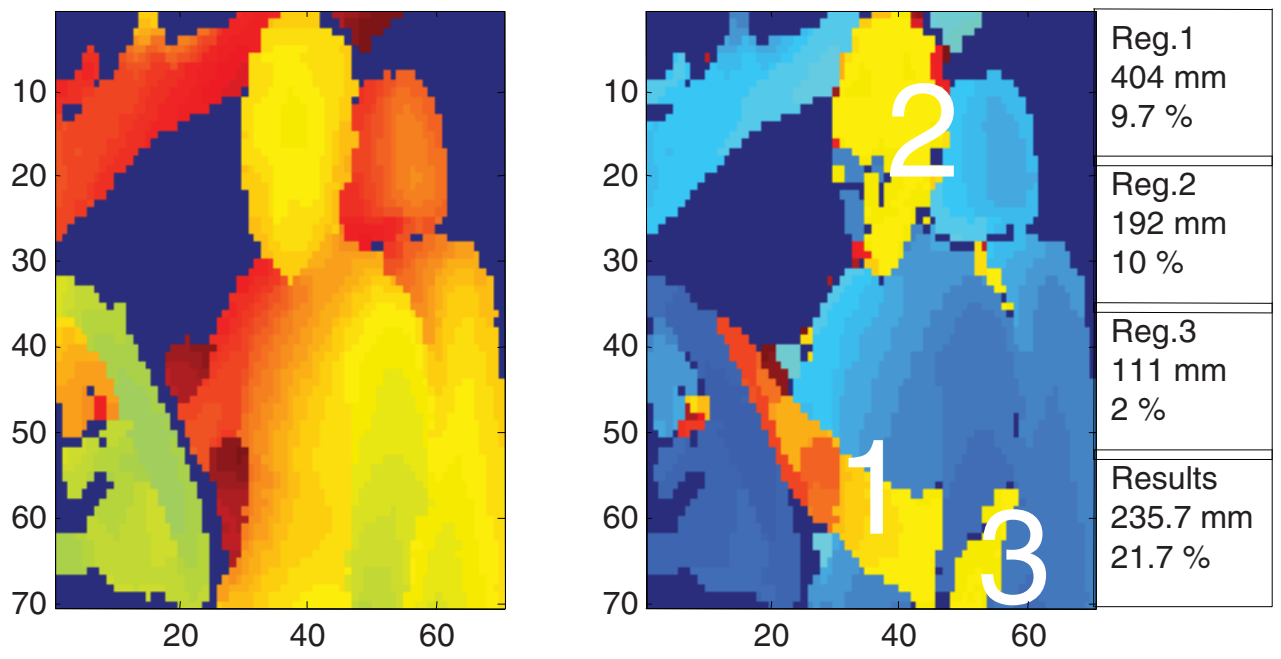

(g) (h)

Figure 6. Motion analyzing in lower resolution
(a and b) Calling while driving
(c and d) Drinking while driving
(e and f) Texting while driving
(g and h) Manipulating vehicle instruments while driving

## C. *A failed detection*

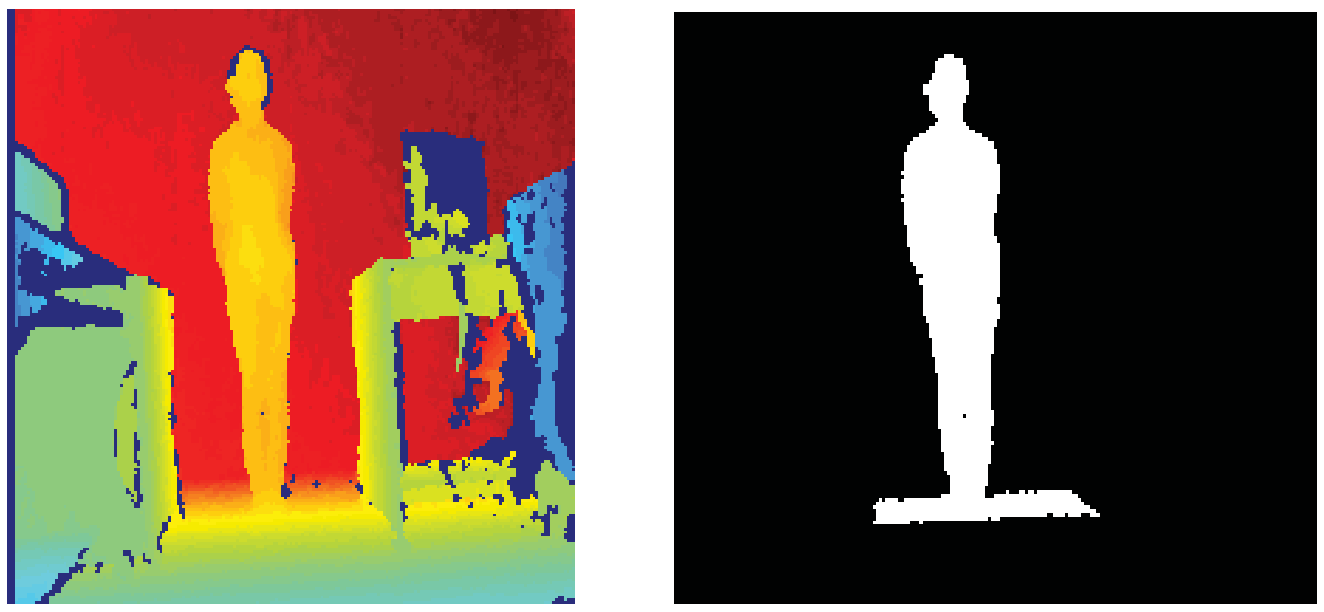

(a) (b)

Figure 7. Human on ground
(a) Original depth image
(b) Extracted result with advanced region growing algorithm

In order to compare our algorithm with the one in [7] in term of separating human feet and local ground plane, we established an experiment with a human standing on the floor as shown in fig. 7a. Unfortunately, our approach could not handle this challenge either (fig. 7b). The fundamental cause of this failure is the obtained depth values of both human feet and local ground plane belonged to a same depth layer. In this case, the best approach should be plane detection method such as RANSAC plane fitting [11] or V-disparity image [12], [13].

## V. CONCLUSION

Our major contribution in this study is a segmentation framework based on the histogram of depth and an adaptive region growing process. Advantages of the proposed method are twofold. The use of depth interval threshold enables the segmentation of multiple human objects. In case of more than one object appeared in the same depth interval, the region growing ensures the accurate separation of objects without prior knowledge of distance between them. This result is very encouraging as it shows the ability of adaptation to different environment conditions, which is critical in the segmentation area.

On the other hand, some limitations of this study still persist that need further investigation. Firstly, the region growing algorithm could not separate the human feet and the local ground plane due to the lack of depth difference between them (Fig. 7). Secondly, the human detector which strongly depends on the geometric structure could misidentify other objects with the similar size and shape. In future work, machine learning will be applied to extract more human features to dealing with those limitations.

## ACKNOWLEDGEMENT

This work was supported by Vietnam National Foundation for Science and Technology Development (NAFOSTED).